\documentclass[runningheads]{llncs}

\usepackage{graphicx}
\usepackage{cite}
\usepackage{algorithmic}
\usepackage{graphicx}
\usepackage{textcomp}
\usepackage{xcolor}
\usepackage{graphicx}
\usepackage{amsmath}
\usepackage{amssymb}
\usepackage{amsfonts}
\usepackage{hyperref}
\usepackage[detect-all=true]{siunitx}
\usepackage{blkarray}
\usepackage{todonotes}
\usepackage{cleveref}
\usepackage{makecell}
\usepackage{multirow}
\usepackage{booktabs}
\usepackage{tabularx}
\usepackage{arydshln}
\usepackage{xspace}
\usepackage{calc}
\usepackage{caption}
\usepackage{subcaption}
\usepackage[normalem]{ulem}
\usepackage[list=true, font=large, labelfont=bf, 
labelformat=brace, position=top]{subcaption}

\newif\ifappendix
\appendixtrue

\newcommand{\ce}{Categorical Cross Entropy\xspace}
\newcommand{\shortce}{CCE\xspace}

\newcommand{\sbce}{Similarity Based Loss\xspace}
\newcommand{\shortsbce}{SimLoss\xspace}

\newcommand{\signbetter}[1]{#1\(^+\)\hspace*{-0.7em}}
\newcommand{\signworse}[1]{#1\(^-\)\hspace*{-0.7em}}

\newcommand{\para}[1]{\noindent\normalfont\textbf{#1}}

\def\signed #1{{\leavevmode\unskip\nobreak\hfil\penalty50\hskip2em
  \hbox{}\nobreak\hfil(#1)%
  \parfillskip=0pt \finalhyphendemerits=0 \endgraf}}

\newsavebox\mybox
\newenvironment{aquote}[1]
  {\savebox\mybox{#1}\begin{quote}}
  {\signed{\usebox\mybox}\end{quote}}

\allowdisplaybreaks

\begin{document}

\title{\shortsbce: Class Similarities in Cross Entropy}
\titlerunning{\shortsbce}

\author{Konstantin Kobs \and
Michael Steininger \and
Albin Zehe \and
Florian Lautenschlager 
\and Andreas Hotho
}
\authorrunning{K. Kobs et al.}

\institute{Julius-Maximilians University W\"urzburg
\email{\{kobs,steininger,zehe,lautenschlager,hotho\}@informatik.uni-wuerzburg.de}
}
\maketitle              
\begin{abstract}

One common loss function in neural network classification tasks is \ce (\shortce), which punishes all misclassifications equally.
However, classes often have an inherent structure.
For instance, classifying an image of a rose as ``violet'' is better than as ``truck''.
We introduce \shortsbce, a drop-in replacement for \shortce that incorporates class similarities along with two techniques to construct such matrices from task-specific knowledge.
We test \shortsbce on Age Estimation and Image Classification and find that it brings significant improvements over \shortce on several metrics.
\shortsbce therefore allows for explicit modeling of background knowledge by simply exchanging the loss function, while keeping the neural network architecture the same.\footnote{Code and additional resources: \url{https://github.com/konstantinkobs/SimLoss}.}

\end{abstract}

\begin{aquote}{Common proverb}
	\textit{Roses are red, violets are blue,\\both are somehow similar, but the classifier has no clue.}
\end{aquote}

\section{Introduction}

\begin{figure*}[t]
	\center
	\includegraphics[height=2.5cm]{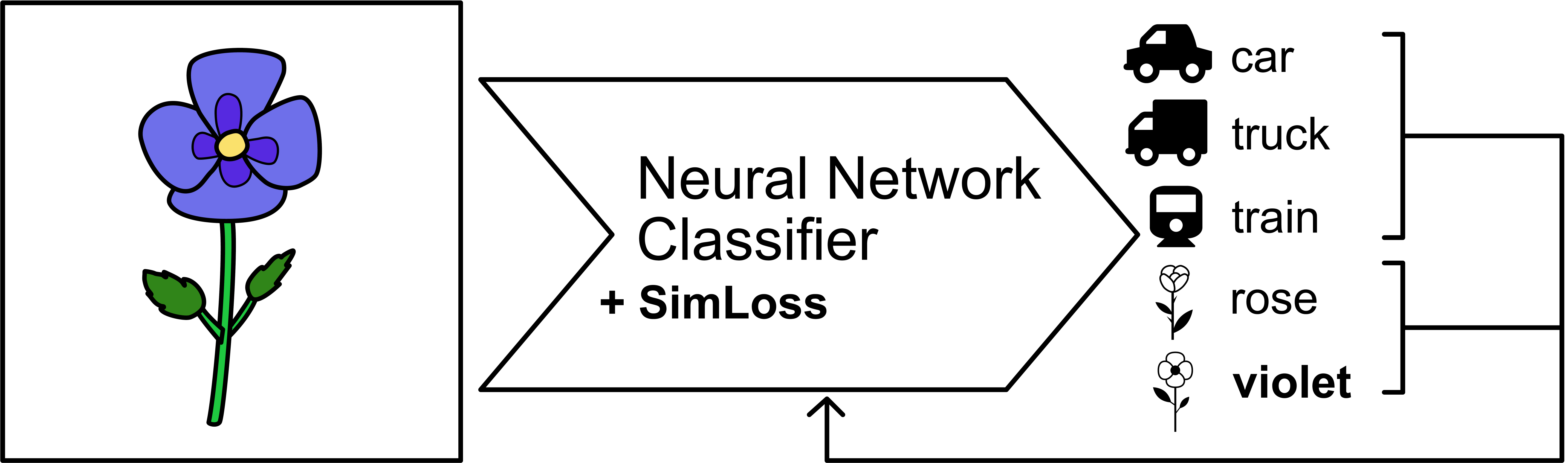}
	\caption{\shortsbce includes knowledge about class relations in the loss function.}
	\label{fig:figure1}
\end{figure*}

One common loss function in neural network classifiers is \ce (\shortce).
\shortce tries to maximize the assigned target class probability and punishes every misclassification in the same way, independent of other information about the predicted class.
Often, however, classes have a special order or are similar to each other, such as different flowers in image classification.
Including class similarities using the inherent class structure (e.g., class order), class properties (e.g., class names) or external information about the classes (e.g., knowledge graphs) in the training procedure would allow the classifier to make less severe mistakes as it learns to predict similar classes.

In this work, we modify \ce and propose \sbce (\shortsbce) as a way to explicitly introduce background knowledge into the training process, as visualized in \Cref{fig:figure1}.
For this, we augment \shortce with a matrix containing class similarities and propose two techniques in order to prepare such matrices that exploit certain class relations: class order and general class similarities.
We show on two tasks, Age Estimation (exploiting class order) and Image Classification (exploiting semantic similarities using word embeddings), that \shortsbce can significantly outperform \shortce.
We also show that tuning the hyper-parameters of both generation techniques influences the model's performance on metrics measuring either more or less specific predictions.

Our contribution is twofold:
First, we introduce a drop-in replacement for \shortce that incorporates class similarities to support the training of neural network classifiers.
Second, we describe two techniques to convert task-specific knowledge into matrices that can be used in the proposed loss function.

\section{Related Work}
\label{sec:related_work}

Previous work on including task-specific knowledge in classification is mostly designed for specific use cases, requires modifications to the model architecture or training procedure, or implicitly learns the information while training.
Sukhbaatar et al. \textit{implicitly} learn a probability instead of a similarity matrix (we provide an analysis of the relationship in the online material) that indicates the chance of a falsely assigned class label in order to compensate for noise~\cite{sukhbaatar2014training}.
This, however, requires changes in the network architecture and a special training procedure.
An analysis of the relation between probability and similarity based matrices is further analyzed in the online resources.
Related to tasks with similar classes are tasks where classes have a taxonomic structure, which is called hierarchical classification.
Specifically designed loss functions and/or model architectures use the fact that classes that belong to the same category are more similar than others~\cite{cesa2006incremental,wu2017hierarchical}.
Izbicki et al. exploit the geospatial relation between areas on earth to automatically geotag input photos~\cite{izbicki2019exploiting}.
Their model learns to predict a mixture of densities that spread across multiple areas instead of specific classes/areas.
A number of task-specific methods try to use the inherent class order of so-called ordinal classification tasks~\cite{fu2008human,guo2009human}.
For example, Niu et al. use multiple binary classifications each indicating whether the value is greater than the class value~\cite{niu2016ordinal}.
Model architectures incorporating semantic similarities using word embeddings were shown to usually predict more similar classes if they fail compared to models without similarity information~\cite{frome2013devise,norouzi2013zero}.
In contrast to the related work, the use of \shortsbce does not require special model architectures and works on any common neural network classifier.
This makes it easy to explicitly support the training procedure with background knowledge.

\section{\sbce}
\label{sec:methodology}

Our proposed \sbce (\shortsbce) is based on the \ce (\shortce).
\shortce assumes that only one class is correct and is defined as \(L_{\text{\tiny\shortce}} = - \frac{1}{N} \sum_{i=1}^{N} \log(\vec{p}_i[y_i])\),
where \(N\) is the size of the dataset and \(\vec{p}_i[y_i]\) is the probability vector output of the network at the target index \(y_i\) for the \(i\)th example.
To model additional knowledge, \shortsbce adds a matrix $\vec{S}$, which gives
\begin{equation}
	\label{eq:sbce}
	L_{\text{\tiny\shortsbce}} = - \frac{1}{N} \sum_{i=1}^{N} \log\left( \sum_{c=1}^{C} \vec{S}_{y_i, c} \cdot \vec{p}_i[c] \right),
\end{equation}
where \(\vec{S} \in [0, 1]^{C \times C}\) encodes class relations.
\(\vec{S}_{i, j}\) is the similarity between classes  \(i\) and \(j\).
\(\vec{S}_{i, j} = 1\) if and only if classes $i$ and $j$ are identical or interchangeable.

\shortsbce is equal to \shortce if \(\vec{S} = I_c\) (identity matrix).
Non zero values lead to smaller losses when the network gives a high score to classes similar to the correct one.
For misclassifications, this leads the network to predict similar classes.

\para{Matrix Generation}
\label{sec:matrix_generation}
We now propose two techniques to generate the matrix \(\vec{S}\), which explicitly captures background knowledge about class relations.
Our techniques allow the modeling of class order and general class similarities.

\textit{Class Order:}
\label{sec:matrix_ord_class}
If classes have an inherent order, we can calculate class similarities based on the distance between the class indices.
As classes lying next to each other are more similar, we construct the similarity matrix $\vec{S}$ as follows:
Assuming the same distance between neighboring classes, we define the reduction factor $r \in [0, 1)$ to be the rate at which the similarity will get smaller given the distance to the correct class.
The similarity matrix is then
\begin{equation}
	\label{eq:matrix_ord_class}
	\vec{S}_{i,j} = r^{|i-j|} \quad \forall i, j \in \{1, \dots, C\}.
\end{equation}
The smaller the reduction factor, the faster the entries converge to \num{0} with increasing distance to the target class.
If the reduction factor is set to \num{0}, the matrix becomes the identity, resulting in the \shortce loss.
The reduction factor is a hyper-parameter of this technique, which can be tuned using a validation dataset to optimize the model for different metrics, as we show in \Cref{sec:experiments}.
As \shortsbce is equivalent to \shortce when $r=0$ (assuming \(0^0 = 1\)), an optimized $r$ will always perform at least as good as \shortce unless we overfit.

\textit{General Class Similarity:}
\label{sec:matrix_img_class}
For some classification tasks, a similarity between classes, such as class names, is available or can be defined.
Then, we can use an appropriate similarity measure $sim: C \times C \rightarrow [0, 1]$ that returns the similarity for two classes \(i, j \in \{1, \dots, C\}\) and calculate all entries of the similarity matrix \(\vec{S}\).
Such similarity measures can be manual, semi- or fully-automatic.
Additionally, we define a lower bound \(l \in [0, 1)\) as a hyper-parameter that controls the minimal class similarity that should have an impact on the network punishment.
We cut all similarities below \(l\) and then scale them such that \(l\) becomes \num{0}:
\begin{equation}
	\label{eq:matrix_img_class}
	\vec{S}_{i,j} = \frac{max(0, sim(i, j) - l)}{1 - l} ~ \forall i,j \in \{1,...,C\} .
\end{equation}
Assuming only the diagonal of \(\vec{S}\) are ones, converging \(l \rightarrow 1\) leads to the \shortce loss, as only the ones in the diagonal are preserved by the lower bound cut-off.

\section{Experiments}
\label{sec:experiments}
In the following, we compare \shortsbce to \shortce by applying them to the same neural network model with the same hyper-parameters
for Age Estimation and Image Classification.
\textit{Age Estimation} is an ordinal classification task with the goal of predicting the age of a person given an image of their face.
The classes have an inherent order: two classes are more similar if they represent similar ages.
A misclassification is thus less harmful for nearer classes.
In \textit{Image Classification}, the goal is to recognize an object shown in an image.
Here, we use class name word embeddings to model semantic class similarities.
For example, classifying an image of a rose as ``violet'' is less harmful than classifying it as ``truck''.

\para{Datasets and Resources}
\label{sec:datasets}
For \textit{Age Estimation}, we train neural networks on the UTKFace \cite{zhang2017age} and AFAD \cite{niu2016ordinal} datasets, both containing human face images annotated with their age.
For UTKFace, we use all images for ages \num{1} to \num{90}, while AFAD has 61 age classes.
We randomly sample training/validation/test sets using 60/20/20 splits.
For \textit{Image Classification}, we use the CIFAR-100 dataset~\cite{krizhevsky2009learning}.
We also use word embeddings from a word2vec model pretrained on Google News~\cite{mikolov2013efficient} to calculate the semantic similarity between class names.
Four class names do not yield a word embedding and are therefore eliminated.
Each remaining class has \num{450} training, \num{50} validation, and \num{100} test examples.

\para{Evaluation Metrics}
\label{sec:evaluation_metrics}
To evaluate our method, we employ task-dependent evaluation metrics that focus both on correct predictions and the similarity of predicted and target class.
For \textit{Age Estimation:}
Accuracy (Acc), Mean Absolute Error (MAE), and Mean Squared Error (MSE).
Accuracy captures exact predictions, while MAE and MSE capture the distance to the target class, thus considering class order.
\textit{Image Classification:}
Accuracy, Superclass Accuracy (SA), and Failed Superclass Accuracy (FSA).
Every example in the CIFAR-100 dataset has a main class and a superclass (e.g., classes ``rose'' and ``orchid'' have the superclass ``flower'').
Superclass Accuracy is the fraction of examples that are correctly put into the corresponding superclass.
This value is always at least as high as Accuracy, as a correctly assigned class implies the correct superclass.
Failed Superclass Accuracy only observes misclassified examples, thus measuring the similarity of misclassifications compared to the target class.
A high FSA means that if the model predicts the wrong class, the predicted class is at least similar to the correct class.
Accuracy only counts exact predictions, while SA and FSA focus on the semantic similarity of the prediction to the target class.

\para{Generating the Similarity Matrix}
\label{sec:exp_matrices}
Since \textit{Age Estimation} has equidistant classes, the similarity matrix can be built using \Cref{eq:matrix_ord_class} without any modifications.
In \textit{Image Classification}, we define the similarity matrix as the cosine similarity $sim_{cos}: w \to [-1,1]$ between class name embeddings, where $sim_{cos}(w,w) = 1$.
To ensure compatibility with the definition in \Cref{sec:matrix_img_class}, we set $sim(i,j) = \max(0,sim_{cos}(w_i,w_j))$ in ~\Cref{eq:matrix_img_class}.

\para{Experimental Setup}
\label{sec:experimental_setup}
Since \shortsbce is a drop-in replacement for \shortce, we investigate the effects of changing the loss function on our example tasks.
Recall that we do not focus on task specific models, but rather on the evaluation of \shortsbce as a general loss function which can be used on various tasks.
Both classification tasks are typical examples for using \shortce.
For \textit{Age Estimation}, we take the Convolutional Neural Network (CNN) from~\cite{niu2016ordinal} and change the output size to be the dataset's number of classes.
The input images are resized to \SI{60}{px} by \SI{60}{px} and the values of all color channels are standardized.
We use the softmax function and apply the \shortsbce loss function using the similarity matrix introduced above.
We study the effect of the reduction factor \(r\) by performing grid search for \(r \in \{0.0, 0.1, \dots, 0.9\}\) on the validation set.
Optimizing the network using Adam~\cite{kingma2014adam} with a learning rate of \num{0.001} and a batch size of \num{1024}, we employ early stopping~\cite{morgan1990generalization} with a patience of \num{10} epochs on the validation MAE.
We smooth random differences (e.g., by weight initialization) by averaging over \num{10} runs.
For \textit{Image Classification}, the LeNet CNN~\cite{lecun1998gradient} is used.
Global standardization is applied to the color channels of the input images.
We stop early if the Accuracy on the validation set plateaus for \num{20} epochs of the Adam optimizer with a learning rate of \num{0.001}, and a batch size of \num{1024}.
We optimize the matrix generation technique's lower bound $l \in \{0.0, 0.1, \dots, 0.8, 0.9, 0.99\}$ with grid search and average \num{10} runs per configuration.
\(l=0.99\) makes the loss equivalent to \shortce, cutting all similarities except the diagonal.

\section{Results}
\label{sec:results}

\Cref{tab:ord_class_results} shows the resulting mean metrics for the validation and test sets given a reduction factor $r$ for both \textit{Age Estimation} datasets.
The best performing reduction factors on the validation and test set are always higher than \num{0.0}, meaning that \shortsbce outperforms \shortce.
Choosing the reduction factor then depends on the metric to optimize for.
For UTKFace, a reduction factor of \num{0.3} leads to the best validation Accuracy, while \num{0.8} or \num{0.9} optimize MAE and MSE, respectively.
For AFAD, \(r=0.5\) yields the best validation result on Accuracy, while \(r = 0.7\) results in the best MAE and MSE.
Overall, choosing a smaller reduction factor \(r \approx 0.4\) optimizes the Accuracy, while larger \(r \approx 0.8\) optimizes MAE and MSE.
This is because large \(r\) lead to higher matrix values and thus smaller punishments for estimating a class near the correct age.
A model optimized for that is favored by metrics that accept approximate matches, such as MAE or MSE.

\begin{table}[t]
    \center
    \caption{Validation and test results averaged over 10 runs on UTKFace and AFAD. Accuracy (Acc) is given in percent. Best validation values are written in bold. Statistically significantly different test values are marked by \(+\) or \(-\), if they are on average better or worse than \shortce (i.e. \(r=0.0\)).}
    \begin{tabular}{@{}c@{\hskip 5pt}c@{\hskip 2pt}c@{\hskip 1pt}c@{\hskip 1pt}c@{\hskip 1pt}c@{\hskip 5pt}c@{\hskip 5pt}c@{\hskip 3pt}c@{\hskip 1pt}c@{\hskip 2pt}c@{\hskip 2pt}c@{\hskip 1pt}c@{\hskip 1pt}c@{\hskip 5pt}c@{\hskip 5pt}c@{\hskip 3pt}}
    \toprule
     & \multicolumn{7}{c}{UTKFace} && \multicolumn{7}{c}{AFAD} \\
    \cmidrule{2-8} \cmidrule{10-16} \multirow{2}*{$r$} & \multicolumn{3}{c}{Validation} & \phantom{a} & \multicolumn{3}{c}{Test} & \phantom{a} & \multicolumn{3}{c}{Validation} & \phantom{a} & \multicolumn{3}{c}{Test} \\
    \cmidrule{2-4} \cmidrule{6-8} \cmidrule{10-12} \cmidrule{14-16} & Acc & MAE & MSE && Acc & MAE & MSE && Acc & MAE & MSE && Acc & MAE & MSE \\
    \midrule
    0.0 & 15.23          &  7.09 			& 122.12          && 14.47             &  7.39             &  131.65  && 11.17          & 4.05          & 32.61          && 11.22             & 4.10             &  33.64            \\
    \midrule
    0.1 & 15.43          &  7.06 			& 119.87          && 14.48             &  7.29             &  127.18  && 11.21          & 4.06          & 32.75          && 11.30             & 4.10             &  33.73            \\
    0.2 & 15.94         &  7.06 			& 121.28          && 14.57             &  7.27             &  127.13   && 11.40          & 4.09          & 33.52          && 11.37             & \signworse{4.15} &  \signworse{34.60}           \\
    0.3 & \textbf{16.25} &  6.95 			& 117.67          && \signbetter{15.17} &  \signbetter{7.19} &  125.70  && 11.34          & 4.10          & 33.53          && \signbetter{11.38} & \signworse{4.16} &  \signworse{34.53}            \\
    0.4 & 16.13         &  6.95 			& 117.52          && \signbetter{15.46} &  \signbetter{7.18} &  125.74  && 11.33          & 4.10          & 33.44          && \signbetter{11.45} & \signworse{4.16} &  \signworse{34.56}            \\
    0.5 & 16.10          &  6.89 			& 115.59          && 15.09 		    &  \signbetter{7.18} &  123.94  && \textbf{11.44} & 4.06          & 33.02          && \signbetter{11.49} & 4.13             &  34.21            \\
    0.6 & 15.62          &  6.83 			& 112.85          && 14.34 		    &  \signbetter{7.09} &  \signbetter{120.34}  && 11.26          & 4.01          & 31.99          && 11.31             & \signbetter{4.05} &  \signbetter{32.84} \\
    0.7 & 14.39          &  6.79 			& 110.12          && 13.07 		    &  \signbetter{7.08} &  \signbetter{121.19}  && 11.22          & \textbf{3.95} & \textbf{31.17} && 11.11             & \signbetter{4.02} &  \signbetter{32.36} \\
    0.8 & 13.50          &  \textbf{6.74} 	& 108.80          && \signworse{12.57} &  \signbetter{7.01} &  \signbetter{117.99}  &&  8.58          & 4.58          & 38.69          && \signworse{8.55}  & 4.64             &  39.78 \\
    0.9 & 9.69          &  6.90 			& \textbf{106.23} && \signworse{9.16}  &  \signbetter{7.18} &  \signbetter{117.62}  &&  6.55          & 5.09          & 44.87          && \signworse{6.47}  & \signworse{5.15} &  \signworse{45.82} \\
    \bottomrule
    \end{tabular}
    \label{tab:ord_class_results}
\end{table}

A Wilcoxon-Signed-Rank-Test with a confidence interval of \SI{5}{\%} shows that optimizing the reduction factor always leads to significant improvements over \shortce.
Sometimes, however, choosing the reduction factor based on a specific metric also results in a trade-off between the chosen and other metrics.

\begin{table}
\center
\caption{Validation and test results over 10 runs with early stopping on the modified CIFAR-100 dataset. Best validation values are written in bold. Statistically significantly different test values are marked by \(+\) or \(-\), if they are on average better or worse than \shortce (i.e. \(l = 0.99\)).}
\begin{tabular}{@{}c@{\hskip 9pt}c@{\hskip 9pt}c@{\hskip 9pt}c@{\hskip 3pt}c@{\hskip 3pt}c@{\hskip 9pt}c@{\hskip 9pt}c@{\hskip 6pt}}
\toprule
\multirow{2}*{$l$} & \multicolumn{3}{c}{Validation} & \phantom{a} & \multicolumn{3}{c}{Test} \\
\cmidrule{2-4} \cmidrule{6-8} & Accuracy & SA & FSA && Accuracy & SA & FSA \\
\midrule
0.99 &         \SI{46.89}{\%}  &         \SI{55.78}{\%}  &         \SI{16.73}{\%}  && \SI{39.51}{\%} & \SI{49.22}{\%} & \SI{16.05}{\%} \\
\midrule
0.90 & \textbf{\SI{47.42}{\%}} &         \SI{56.32}{\%}  &         \SI{16.95}{\%}  && \signbetter{\SI{40.15}{\%}} & \signbetter{\SI{49.93}{\%}} & \SI{16.36}{\%} \\
0.80 &         \SI{46.37}{\%}  &         \SI{55.38}{\%}  &         \SI{16.80}{\%}  && \SI{39.49}{\%} & \SI{49.32}{\%} & \SI{16.22}{\%} \\
0.70 &         \SI{46.95}{\%}  &         \SI{55.92}{\%}  &         \SI{16.90}{\%}  && \SI{39.86}{\%} & \SI{49.63}{\%} & \SI{16.25}{\%} \\
0.60 &         \SI{47.28}{\%}  & \textbf{\SI{56.44}{\%}} &         \SI{17.39}{\%}  && \SI{40.00}{\%} & \SI{50.00}{\%} & \signbetter{\SI{16.67}{\%}} \\
0.50 &         \SI{46.36}{\%}  &         \SI{56.18}{\%}  &         \SI{18.28}{\%}  && \SI{39.26}{\%} & \SI{49.40}{\%} & \signbetter{\SI{16.70}{\%}} \\
0.40 &         \SI{38.03}{\%}  &         \SI{50.58}{\%}  &         \SI{20.28}{\%}  && \signworse{\SI{32.18}{\%}} & \signworse{\SI{44.58}{\%}} & \signbetter{\SI{18.30}{\%}} \\
0.30 &         \SI{28.65}{\%}  &         \SI{43.76}{\%}  & \textbf{\SI{21.18}{\%}} && \signworse{\SI{24.43}{\%}} & \signworse{\SI{38.90}{\%}} & \signbetter{\SI{19.13}{\%}} \\
0.20 &         \SI{21.66}{\%}  &         \SI{37.97}{\%}  &         \SI{20.80}{\%}  && \signworse{\SI{18.54}{\%}} & \signworse{\SI{33.68}{\%}} & \signbetter{\SI{18.58}{\%}} \\
0.10 &         \SI{16.40}{\%}  &         \SI{31.68}{\%}  &         \SI{18.31}{\%}  && \signworse{\SI{14.25}{\%}} & \signworse{\SI{28.70}{\%}} & \SI{16.85}{\%} \\
0.00 &         \SI{ 2.80}{\%}  &         \SI{ 8.37}{\%}  &         \SI{ 5.77}{\%}  && \signworse{\SI{ 2.53}{\%}} & \signworse{\SI{ 8.06}{\%}} & \signworse{\SI{ 5.71}{\%}} \\
\bottomrule
\end{tabular}
\label{tab:img_class}
\end{table}

For the \textit{Image Classification} task, \Cref{tab:img_class} shows the results for the validation and test set of the CIFAR-100 dataset given a lower bound $l$.
On average, the best performing model always has a lower bound of less than \num{0.99}, again showing that \shortsbce outperforms \shortce.
Also, a statistical test reveals that \(l=0.9\) gives significantly better results on the test set in terms of Accuracy and Superclass Accuracy.
Smaller lower bounds tend to reduce the Accuracy as the loss function hardly punishes any misclassification.
For $l \approx 1$, the loss is equivalent to \shortce, forcing the network to predict the correct class, thus increasing Accuracy.
In between, the network is guided to predict the correct class but is also not punished severely for misclassifications of similar classes.
This improves Superclass Accuracy, which pays attention to more similar classes.

\begin{figure}[b]
\begin{subfigure}[t]{0.49\textwidth}
	\includegraphics[width=\textwidth]{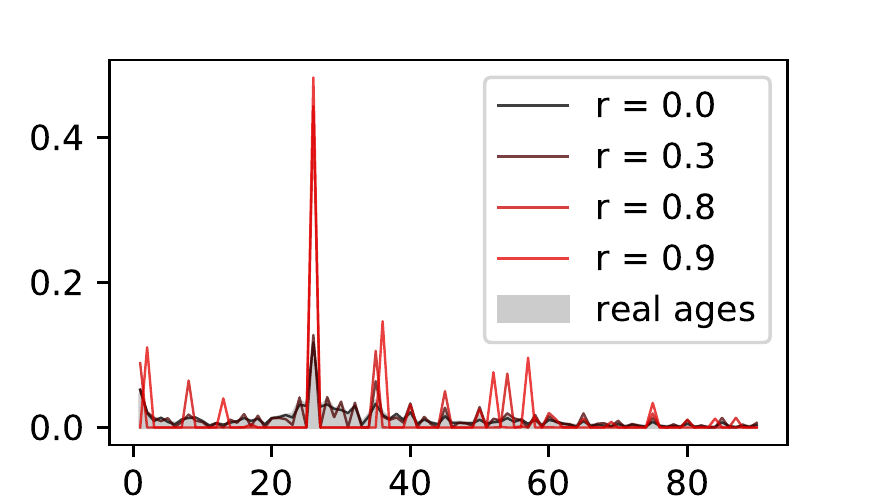}
	\subcaption{All examples. \shortce fits the real data distribution the best.}
	\label{fig:mean_output_all}
\end{subfigure}
\hfill
\begin{subfigure}[t]{0.49\textwidth}
	\includegraphics[width=\textwidth]{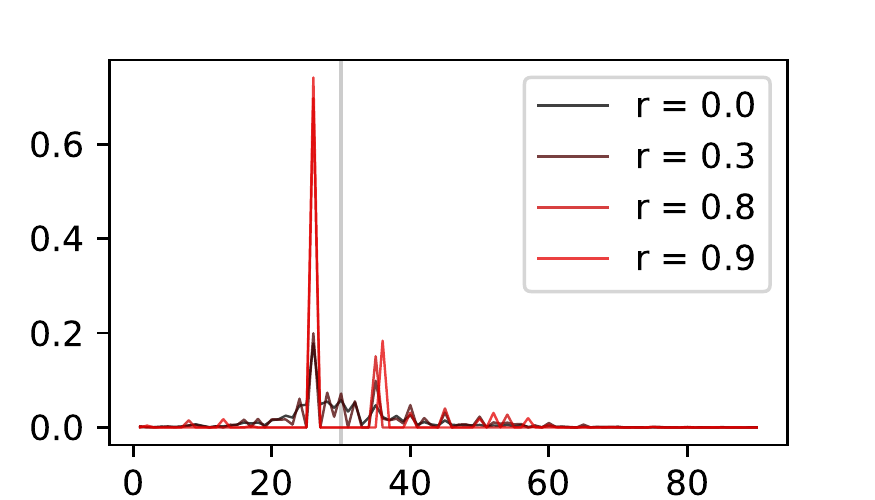}
	\subcaption{All examples of class ``30''. The grey line indicates the target age.}
	\label{fig:mean_output_age}
\end{subfigure}
\caption{Mean probability distribution output for different \(r\). High reduction factors lead the network to choose only few representative classes.}
\end{figure}

\para{Analysis}
\label{sec:analysis}
To understand the effect of \shortsbce, we focus on Age Estimation whose one dimensional classes are easy to visualize.
We compare the best models for UTKFace trained using \shortsbce and \shortce, i.e. $r \in \{0.0,\allowbreak 0.3,\allowbreak 0.8,\allowbreak 0.9\}$.
For each $r$, we plot the mean output distribution for all examples in the dataset as well as the real age distribution, which is shown in \Cref{fig:mean_output_all}.
\shortce ($r = 0.0$) resembles the real age distribution the best, while higher reduction factors tend to aggregate groups of multiple age classes.
With a higher reduction factor, the number of spikes decreases and the distances between them increase:
The model chooses representative classes to which it mainly distributes the output probability mass.
This becomes apparent in \Cref{fig:mean_output_age}, where we plot the mean output distribution for all examples of age \num{30}.
The network with \(r = 0.9\) focuses its probability output to the two nearest representative classes, in this case ``26'' and ``35''.
The Accuracy of the network decreases, as the output probability mass is not on the correct class, but the distance of the prediction to the correct class is smaller than for \shortce.
Representative classes are apparently chosen such that frequent items receive more probability mass from the model.
A higher reduction factor therefore leads to a coarser class selection.
This can be explained by the optimization objective of the loss function.
The loss should be smaller for misclassifications of similar classes than for dissimilar classes.
Representing multiple similar classes as one class and predicting it more often for similar classes does not lead to the smallest possible loss value.
However, the loss gets smaller compared to predicting dissimilar classes, as the punishment should be smaller for classifying a similar class.
In the case of Age Estimation, predicting an age that lies close to the correct age will decrease the Accuracy, but perform better than \shortce on MAE and MSE.
In Image Classification, selecting one or multiple representative classes leads to smaller Accuracy but to higher Superclass Accuracy and Failed Superclass Accuracy than \shortce.
Higher similarities in the matrix thus guide the network to make coarser predictions, improving metrics that accept predictions of similar classes.
The results from \Cref{sec:experiments} also show that keeping the loss near \shortce by choosing the similarity matrix conservatively can improve on specific prediction metrics such as Accuracy as well.

\section{Conclusion}
\label{sec:conclusion}

In this work, we have presented \shortsbce, a modified \ce loss function that incorporates background knowledge about class relations in form of class similarities.
We have introduced two techniques to prepare similarity matrices to exploit class order and general class similarity that can be used to significantly improve the performance of neural network classifiers on different metrics.
Also, \shortsbce helped with predicting more similar classes if the model misclassified an example.
In our analysis, we found that \shortsbce forced the model to focus on choosing representative classes.
The number of representative classes can be implicitly tuned by a hyper-parameter.
While finding the best hyper-parameter and similarity metric can be computationally expensive and non-trivial, \shortsbce can incorporate arbitrary similarity metrics into a classifier.

\bibliographystyle{splncs04}
\bibliography{paper}

\ifappendix

\section*{Appendix}

\subsection*{Relation between Similarity and Probability-based Matrices in \shortsbce}

Some works use a loss function similar to our proposed \shortsbce.
Instead of similarities, the matrix \(\vec{S}\) consists of probabilities, such that each row sums to one~\cite{sukhbaatar2014training,izbicki2019exploiting}.
We will call this loss \(L_{\text{prob}}\).
We discuss the relation between both loss versions --- similarity versus probability matrix --- in this appendix.

We can show that both similarities and probabilities in the matrix lead to the same gradients:
We can transform our loss \(L_{\text{\tiny\shortsbce}}\) into \(L_{\text{prob}}\) by dividing each similarity matrix entry by the row's sum.
This loss depends on the network's output --- the probability distribution \(p\) --- as well as the corresponding target class indices \(y\).
It can be written as:

\begin{align*}
    L_{\text{prob}} &= - \frac{1}{N} \sum_{i=1}^{N} \log\left( \sum_{c=1}^{C} \frac{1}{\sum_{c' = 1}^{C} \vec{S}_{y_i, c'}} \vec{S}_{y_i, c} \cdot \vec{p}_i[c] \right) \\
        &= - \frac{1}{N} \sum_{i=1}^{N} \log\left( \frac{1}{\sum_{c' = 1}^{C} \vec{S}_{y_i, c'}}  \sum_{c=1}^{C} \vec{S}_{y_i, c} \cdot \vec{p}_i[c] \right) \\
        &= - \frac{1}{N} \sum_{i=1}^{N} \left[\log\left( \sum_{c=1}^{C} \vec{S}_{y_i, c} \cdot \vec{p}_i[c] \right) - \log\left( \sum_{c' = 1}^{C} \vec{S}_{y_i, c'} \right) \right] \\
        &= - \frac{1}{N} \sum_{i=1}^{N} \left[\log\left( \sum_{c=1}^{C} \vec{S}_{y_i, c} \cdot \vec{p}_i[c] \right) \right] + \frac{1}{N} \sum_{i=1}^{N} \left[ \log\left( \sum_{c' = 1}^{C} \vec{S}_{y_i, c'} \right) \right] \\
        &= L_{\text{\tiny\shortsbce}} + \frac{1}{N} \sum_{i=1}^{N} \left[ \log\left( \sum_{c' = 1}^{C} \vec{S}_{y_i, c'} \right) \right] \quad . \\
\end{align*}

The second summand leads to different loss function values for both matrices.
It does not depend on the probability output \(p\).
Therefore, when calculating the gradients with respect to $p$, this term becomes zero:

\begin{align*}
    \frac{\partial}{\partial p} L_{\text{\tiny\shortsbce}}' &= \frac{\partial}{\partial p} L_{\text{\tiny\shortsbce}} \quad .
\end{align*}

Both \(L_{\text{\tiny\shortsbce}}\) and \(L_{\text{prob}}\) yield the same gradients when optimizing the model.
If the largest values in the matrices are on the diagonal, both matrix variants will have the same parameters when reaching the global optimum~\cite{zhang2018generalized}.
Even though both methods theoretically lead to the same results, our method is less restrictive since it does not require a probability distribution per row.
For example, while similarities can be calculated for each class pair independently, a probability distribution needs to be normalized over all values in the row prior to use.
For tasks with a large number of classes, the similarity matrix might not need to be stored but could be calculated on the fly, while probabilities would cause larger computational costs.
Especially on edge devices with very limited memory, this is an advantage of our method.

Another advantage of similarities compared to probabilities is that, because the diagonal of the similarity matrix consists of ones, a value of zero can be reached by the loss function, making the loss value more interpretable.
A loss value of zero always means that the probability mass of the neural network output vector is put into the correct class, even if there are similar classes.
Normalizing such a matrix to ensure probability distributions per row would always yield larger loss values, even if the correct class is predicted with \SI{100}{\percent} probability.
Therefore, \(L_{\text{\tiny\shortsbce}}\) always has a lower bound of zero, which gives an interpretable impression of the training status.
In a probability matrix based loss, such an interpretation is not given as the lower bound of the loss depends on the probability distributions in the matrix rows.
A loss value of zero can only be achieved when using the identity matrix.
This, however, would be equivalent to \ce and would not allow for including background knowledge into the model.

\fi

\end{document}